\documentclass[preprint,12pt]{elsarticle}

 \usepackage{graphicx}
\usepackage{psboxit,pstricks}
\usepackage{epsfig}
\usepackage{pmat}
\usepackage{amssymb}
\usepackage[cmex10]{amsmath}
\usepackage{array}
\usepackage{algorithm}
\usepackage{algpseudocode}
\usepackage{color}

\journal{Phys. Med. Biol.}

\begin{document}

\begin{frontmatter}


 \title{Metastatic Liver Tumor Segmentation from Discriminant Grassmannian Manifolds}

  \author[EPM,JUS]{Samuel Kadoury\corref{cor1}}
  \ead{samuel.kadoury@polymtl.ca}
    \author[EPM]{Eugene Vorontsov}
 \ead{eugene.vorontsov@polymtl.ca}
     \author[UDM]{An Tang}
      \ead{duotango@gmail.com}
 \address[EPM]{Ecole Polytechnique Montr\'eal, Montr\'eal, Qu\'ebec, Canada}
  \address[JUS]{CHU Sainte-Justine Hospital Research Center, Montr\'eal, Qu\'ebec, Canada}
  \address[UDM]{University of Montreal, Dept. Radiology, Montr\'eal, Qu\'ebec, Canada}

 \cortext[cor1]{Corresponding author. Prof. Samuel Kadoury, Ph.D., P.O. Box 3079, Succ. Centre-Ville, Montr\'eal, Qu\'ebec, Canada, H3C3A7.}


\begin{abstract}
 
Early detection, diagnosis and monitoring of liver cancer progression can be achieved with precise delineation of metastatic tumors. However accurate automated segmentation remains challenging due to presence of noise, inhomogeneity and high appearance variability of malignant tissue. In this paper, we propose an unsupervised metastatic liver tumor segmentation framework using a machine learning approach based discriminant Grassmannian manifolds which learns the appearance of tumors with respect to normal tissue. First, the framework learns within-class and between-class similarity distributions from a training set of images to discover the optimal manifold discrimination between normal and pathological tissue in the liver. Second, a conditional optimization scheme computes non-local pairwise as well as pattern-based clique potentials from the manifold subspace to recognize regions with similar labelings and incorporate global consistency in the segmentation process. The proposed framework was validated on clinical database of 43 CT images from patients with metastatic liver cancer. Compared to state-of-the-art methods, our method achieves better performance on two separate datasets of metastatic liver tumors from different clinical sites, yielding an overall mean Dice similarity coefficient of $90.7 \pm 2.4$ in over 50 tumors with an average volume of $27.3$mm$^{3}$.

\end{abstract}


\end{frontmatter}

\section{Introduction}
\label{sec:intro}

The delineation of primary tumors or metastases from diagnostic imaging is a crucial but challenging problem in any clinical applications, such as tumor detection, diagnosis, treatment planning for radiofrequency ablation or surgical interventions, and monitoring of treatment response which is done with predefined criterions. Segmentation of tumors in 3D is a pre-requisite for precise clinical measurements that are done during clinical examinations. One example of such an application in liver oncology is the Response Evaluation Criteria in Solid Tumors (RECIST) that measures the size of solid tumors as the maximum diameter \cite{Eisenhauer09}. Ideally, this metric would be made in a volumetric fashion in 3D, instead of repeating the delimitation on each axial plane. However manual segmentation is not applicable in clinical routine as it takes too much time and lacks reproducibility across raters \cite{Pinto10}. Furthermore, to precisely follow the progression of a tumor throughout longitudinal scans, it remains difficult to identify the accurate boundaries of the tumors to quantify the difference in volume. For these reasons, automated tumor segmentation would represent an ideal alternative if it was efficient, reproducible, and permitted accurate delineation of tumor boundaries and tumor volumetry. Further, clinical utility of automated tumor segmentation would be enhanced if it permitted longitudinal comparison of tumor volumetry in subsequent visits.
 \\
 \\
Different methods were previously presented to achieve semi-automated or automated segmentation of metastases in different organs and pathologies, such as in prostate, lung or for neurological disorders. Approaches using discrete Markov Random Fields (MRFs; \cite{Li09}) and Conditional Random Fields (CRFs; \cite{Lafferty01}) were used to model complex dependencies within sample distributions, offering improved segmentation accuracy compared to independent and identically distributed (i.i.d.) classifiers such as Support Vector Machines (SVM). Other works by Li et al. \cite{Li06} located tumor boundaries uses machine learning to classify regions, while Freiman et al. \cite{Freiman08} used a Bayesian classifier to segment various organs with medical images and subsequently applied morphological operators to correct for segmentation discrepancies. Furthermore, a study comparing three semi-automated methods was presented in \cite{Zhou10} using contrast-enhanced CT. In the study by Smeets et al. \cite{Smeets10}, a supervised statistical pixel classification method was trained from a dataset of liver tumors. The technique then applied a level-set adaptation approach to isolate the region of interest. In \cite{Subbanna13}, a hierarchical probabilistic Gabor filter was used in combination with an MRF segmentation framework to delineate brain tumors in MRI volumes, where a Bayesian framework  provides a coarse probabilistic texture-based segmentation of active and edema regions. But despite significant intra- and inter-rater variabilities and prolonged time for manual segmentation, very few of these automated approaches are currently used in clinical practice. Completely automated methods may be prone to suboptimal precision and not be as robust as semi-supervised approaches. Furthermore, they often require high computational requirements. 
\\
\\
State-of-the-art tumor segmentation frequently combines efficient classification techniques with low level segmentation methods. From such perspective, tumor detection is addressed as a classification problem where one aims at separating normal from diseased tissues at the voxel level, while imposing smoothness criterions in the process. In \cite{Bauer11}, SVM classification using multispectral intensities and textures is combined with hierarchical CRFs to segment brain tumors in MR images. In \cite{Hame12}, a semi-automated lesion segmentation method is presented using  Markov Random Fields and non-parametric distributions which are defined locally to estimate the tumor shape. Alternatively, the use of fully-connected pairwise models can improve the performance over traditional MRF models that are based on a finite number of links. Still, pairwise models remain limited to depict higher-order relationships, which could produce results of limited accuracy when modelling the global appearance of an object. On the other hand, higher-order relationships can incorporate a reliable labelling output within inhomogeneous regions, as shown in \cite{Sabuncu10}, while prior information modelled as co-occurrences between various classes has demonstrated significant promise for segmentation \cite{Vineet12}. This idea has been applied for the classification of pathology using discrete hidden MRFs \cite{Charras-Garrido12}.
\\
\\
A major drawback from these methods is that conditional probabilities are obtained directly from the high-dimensional image space, which is not ideal as they fail to express the underlying representation of the dataset and assimilates all measures to Euclidean distances \cite{Roweis00}. They also assume linear discrimination (SVMs) when in most cases, sets are not linearly separable in the image space. In contrast, manifold learning techniques intrinsically consider the nonlinear distribution of the data, and enable important evaluation of test cases in a learned population by using a mapping distance. Recently, various approaches have used manifolds to discriminate the presence of white matter brain lesions \cite{Kadoury12} or track organ motion or discover regional variations within images \cite{Bhatia12}. In \cite{Gerber10}, manifold learning techniques were used to capture the non-linear variability of brain development from a dataset of MR images. However, techniques such as Laplacian eigenmaps are sensitive to outliers and unable to cope with pathological or abnormal tissues \cite{Yang14}. To the best of our knowledge, prior information captured by a discriminant embedding has yet to be exploited in a fully automated graphical model segmentation framework. Preliminary results demonstrated promise but was validated on a limited dataset and could not properly capture inter-tumor inhomogeneities \cite{Kadoury13}. 
\\
\\
In this paper, we propose a segmentation approach for tumors using discriminant manifold embeddings, where the low-dimensional representation are integrated in a graphical model used for segmentation. The contributions are two-fold. First, a discriminant graph-embedding with Grassmannian kernels is generated from prior data to learn the nonlinear intensity distributions of normal liver tissue and pathological regions. Second, we suggest to employ a recently proposed mean-field CRF inference approach where potentials are computed directly from the low-dimensional embedding, capturing the underlying structure. Unary and pairwise potentials assess the proximity to manifold regions and the dissimilarity between pairs of segments, respectively. Higher-order potentials ensure regional consistency to efficiently discriminate tumors from normal tissue. These potentials are integrated in an existing higher-order conditional random field \cite{Kohli09}. We validated the manifold constrained segmentation method on metastatic liver tumors in CT images, from two separate clinical datasets which included diagnositic images of patients which were subsequently treated by radio frequency ablation. 
\begin{figure}
\begin{center}
\includegraphics[width=1\linewidth]{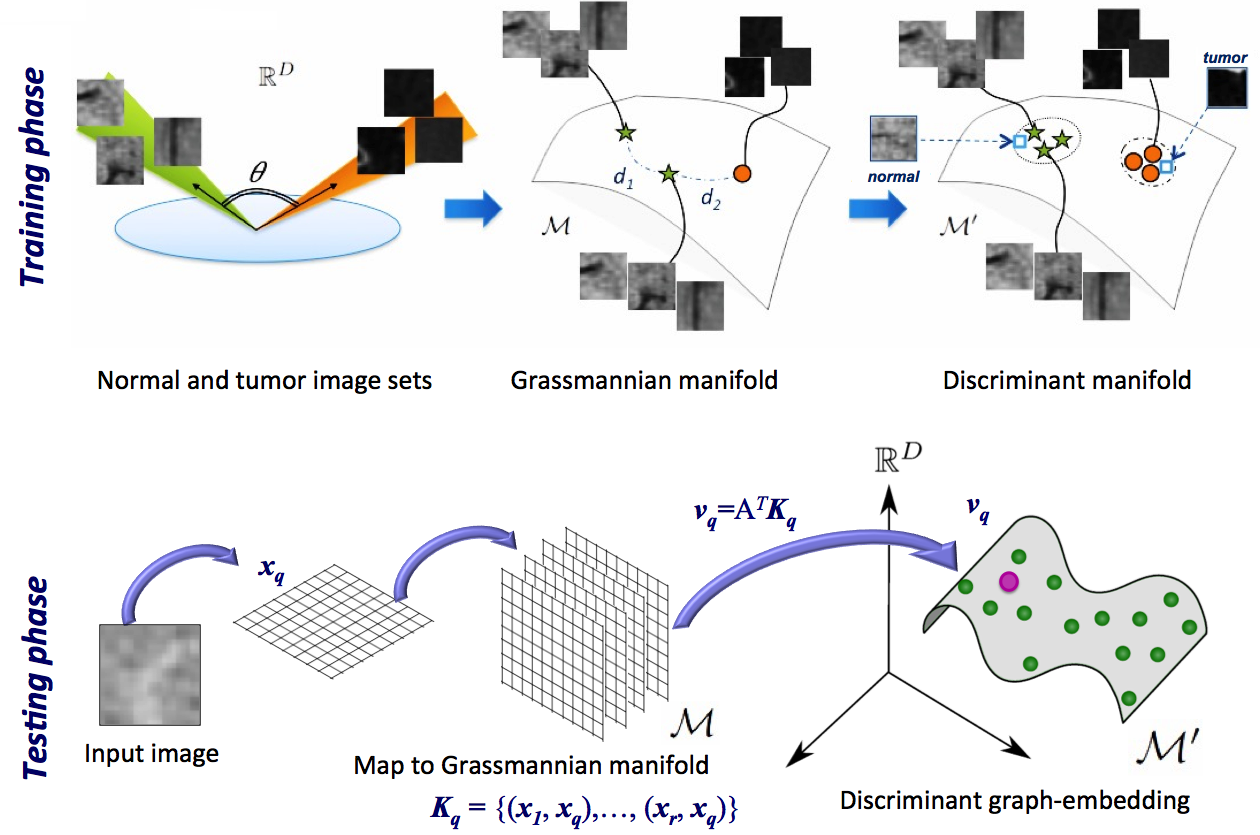}
  \caption{Flowchart diagram of the proposed manifold-based segmentation method used for the extraction of tumors  from medical images. The first phase consists of the training of discriminant Grassmannian manifolds using tumor and normal liver tissue extracted from samples images. The second phase consists of the online segmentation process, with patches extracted from the images and projected onto a Grassmannian manifold which is used by the CRF segmentation process.}
\label{fig:Flow}
\end{center}
\end{figure}
\\
\\
We now describe the proposed framework, with the outline illustrated in Fig. \ref{fig:Flow}. The first phase consists of creating a non-linear, low-dimensional embedding from patches extracted from training datasets of annotated tumor images and liver organ tissue, surrounding the tumor region. These patches are embedded in a Grassmannian manifold, where within and between similarity graphs are created from the manifold points in order to incorporate a discriminatory measure in the learning process. Once the learning phase is completed, a new patient image can be processed using an integrated and interconnected CRF graph to perform the unsupervised  segmentation of tumor regions. This graph involves costs related to manifold support, prior geometrical dependencies and cliques described in the discriminant manifold domain, within regions of interest of the tumor. Pairwise potentials measures the similarity between connected voxels in manifold space, while high-order cliques evaluates the variance of data points within manifold space. The energy function is minimized with primal-dual optimization procedure. A careful selection of the intrinsic dimensionality and parameter settings is performed to properly model the non-linear space.
\\
\\
The paper is structured as follows. The developed method, including the training and segmentation steps is presented in Section 2. The experimental results are reported in Section 3, and Section 4 concludes the paper with a discussion.

\section{Materials and Methods}

In this section, we present the imaging data used in this study and the overall framework of the metastatic liver tumor segmentation. First, the training phase learns the discriminant Grassmannian manifolds using databases of labelled tumor and normal liver tissue in order to determine the mapping function for unseen cases. Then, the segmentation process of a new CT examination uses the manifold potentials in a conditional random field segmentation process to delineate the region of interest.

\subsection{Imaging data}

To train and test the proposed tumor segmentation framework, two distinct CT imaging datasets were used. The first clinical dataset was the public dataset from the 2008 MICCAI segmentation challenge, which was used by several other groups for tumor segmentation \cite{Deng08}. The database includes 4 training CT datasets, which have 10 ground-truth tumor segmentations. The testing database has 13 CT datasets, totalling 20 tumors with ground-truth segmentations. 
\\
\\
The second clinical dataset was composed of a total of 30 images provided by radiology departments from two separate clinical institutions, totalling 40 tumors. The first subset included 12 contrast-enhanced CT examinations, totalling 15 tumors as two patients had multiples tumors. The images were acquired with a 64-slice CT scanner with axial dimensions of 512 by 512, 1mm slice thickness, in-plane resolution of 0.6mm and the number of slices ranging between 112 and 330.  Cases were referred for follow-up or for subsequent treatment such as for chemo-embolization or radio frequency ablation (RFA). The second subset included 18 contrast-enhanced CT examinations acquired also with a 64-slice CT scanner (in-plane resolution of 512 by 512), with a slice thickness varying between 0.9 and 1.2mm, in-plane resolution of 0.8mm and the number of slices ranging between 87 and 274. In this database subset, tumors were limited to liver metastases from colorectal cancer. Tumors sizes varied between 2 and 7 cm in maximum diameter in this database and were manually segmented by an experimented radiologist. Every patient had a single tumor, except 5 patients, who had 3 tumors (2 patients) and 2 tumors (3 patients). The total number of metastatic tumors was 25 tumors. The second dataset was randomly separated into 10 training and 30 testing cases. 

\subsection{Learning Discriminant Grassmannian Manifolds}

Manifold learning algorithms are based on the premise that data are often of artificially high dimension and can be embedded in a lower dimensional space. However the presence of outliers and multi-class information can on the other hand affect the discrimination and/or generalization ability of the manifold. We present here the mechanism to learn the optimal separation between normal and pathological tissue by using a discriminant graph-embedding based on Grassmannian manifolds proposed by \cite{Harandi11}, where it is adapted to the segmentation process. Here, we describe the Grassmannian kernels applied to the input data points, which in turn are embedded into a discriminant manifold domain that incorporates within and between similarity graphs.

\subsubsection{Grassmannian kernels}

In the framework, each sample point $x_i$ (representing an image patch) in a Grassmannian manifold is described by the set of $m$-dimensional domains in $\mathbb{R}^{D}$ ,  described by a orthonormal matrix of $D \times m$. It is then possible to see if two Grassmannian manifold points are similar, by mapping one to another using the $m \times m$ orthogonal matrix. In this work, the similarity between two manifold points ($x_i,x_j$) is measured as a combination of two Grassmannian kernels $\mathbb{K}_{i,j}$ defined in the Hilbert Space such that:
\begin{equation}\label{eq.kernel}
  \mathbb{K}_{i,j} = \alpha_1 k_{i,j}^{[proj]}(x_i,x_j)+ \alpha_2  k_{i,j}^{[CC]}(x_i, x_j)
\end{equation}
with $\alpha_1,\alpha_2 \geq 0 $. The projection kernel defined as  $k_{i,j}^{[proj]}=\| x_i^T x_j\|^2_F$ determines the largest correlation based on the cosine measure between principal vectors of two sets. The second denotes  canonical correlation Grassmanian kernel:
\begin{equation}\label{eq.kernelcc}
   k_{i,j}^{[CC]} = \underset{\textbf{a}_p \in \operatorname{span}(x_i)}{\operatorname{max}}  \, \underset{\textbf{b}_q \in \operatorname{span}(x_j) }{\operatorname{max}}  \textbf{a}^T_p  \textbf{b}_q
\end{equation}
subject to $\textbf{a}^T_p  \textbf{a}_p = \textbf{b}^T_p  \textbf{b}_p =1$ and $\textbf{a}^T_p  \textbf{a}_q = \textbf{b}^T_p  \textbf{b}_q =0, \,\, p \neq q$. This kernel is positive definite since $z^T \mathbb{K}z > 0$ such that $\forall z \in \mathbb{R}^n$ as shown in \cite{Harandi11}, and well-defined since singular values of $x_1^T x_2$ are equal to $\textbf{\emph{R}}_1^T x_1^T x_2 \textbf{\emph{R}}_2$, with $\textbf{\emph{R}}_1,\textbf{\emph{R}}_2 \in Q(o)$, indicating orthonormal matrices of order $o$. This makes the kernel invariant to various representation of the subspaces. While the projection kernel based on principal angles is able to identify the predominant differences from a pair of image sets, the canonical correlation kernel acts as a similarity metric to identify features which offer the best correlation between two sets of images. This later kernel is also invariant to various representations of the subspaces. With each kernel representing different components of the image distributions, the combined kernel  $\mathbb{K}_{i,j}$ is able to cover a wider range of features to efficiently characterize tumors with respect to typical features apparent in normal liver tissue, and identify the discriminant features to separate both classes.

\subsubsection{Graph architecture}

In order to effectively discover the low-dimensional embedding, it is necessary to maintain the local structure of the data in the new embedding. The structure $G=(\textbf{\emph{V}},\textbf{\emph{W}})$ is an undirected similarity graph, using a collection of nodes $\textbf{\emph{V}}$ connected by edges, and a matrix $\textbf{\emph{W}}$ with symmetric values describing  relationships between the nodes. The relationship between the diagonal $\textbf{\emph{D}}$ and the Laplacian $\textbf{\emph{L}}$ matrices is modelled by as $\textbf{\emph{L}}= \textbf{\emph{D}} - \textbf{\emph{W}}$. Here, each element of  $\textbf{\emph{D}}$ describes the sum of the weight matrix $\textbf{\emph{W}}$. By representing an architecture as a set nodes and edges, the goal of discriminant graph embedding is to increase the discriminatory power of embedded data by mapping the high-dimensional distribution data to another low-dimensional space. In the context of embedding, similarities between pairs of vertices remain the same in either space by solving a general eigen-analysis problem \cite{Yan07}.
\\
\\
Here, $N$ labelled points  $\mathbb{X}=\{(x_i,c_i)\}_{i=1}^{N}$ are generated from the underlying manifold $\mathcal{M}$, where $c_i$ denotes the label (tumor or normal). The task at hand is to maximize the discrimination between tumor and normal liver tissue by mapping the underlying data $x_i$ into a vector space $y_i$, while preserving similarities between data points in the high-dimensional space. Discriminant graph-embedding based on locally linear embedding (LLE)  \cite{Roweis00} uses graph-preserving criterions to maintain these similarities, which are included in a sparse and symmetric $N \times N$ matrix, denoted as $M$. The embedding cost function is minimized using the following formulation:
\begin{align}\label{eq.llegraph}
    \sum_{i}\|y_i - \sum_{j} M(i,j) y_j\|^2 &= y^T (I-M)^T (I-M)y\\
     &=\underset{Y^T T=1}{\operatorname{argmin}} \,\,\ y^T(\textbf{\emph{D}}-\textbf{\emph{W}})y .  \nonumber
\end{align}
with $I$ as identity. Therefore, the LLE algorithm is described as a direct graph embedding problem.
\\
\\
\subsubsection{Within and between similarity graphs}

In our work, the manifold embedding  $\mathcal{M}$ is constructed with a within-class graph $\textbf{\emph{W}}_w$, describing the similarity of regions with the same type (tumor and normal liver tissue), and a between-class penalty graph $\textbf{\emph{W}}_b$,  used to discriminate normal and tumor tissue. When constructing the discriminant LLE graph, elements are partitioned between these two classes. The intrinsic graph $G$ is first created by assigning edges only to vertices of the same class (tumor or normal). The local reconstruction coefficient matrix $M(i,j)$ is obtained by minimizing:
\begin{equation}\label{eq.costlle}
\underset{M} \min \sum_{j \in \mathcal{N}_w(i)} \| x_i - M(i,j) x_j \|^2  \,\,\,  \sum_{j \in \mathcal{N}_w(i)}  M(i,j)=1  \,\,\ \forall i
\end{equation}
with $\mathcal{N}_w(i)$ as the neighborhood of size $k_1$, within the same region as point $i$ (e.g. tumor region). Each sample is therefore reconstructed only from images of the same region. The local reconstruction coefficients are incorporated in the within-class similarity graph, such that the matrix $\textbf{\emph{W}}_w$ is defined as:
\begin{equation}\label{eq.simwithin}
W_w(i,j)=
\begin{cases}
(M+M^T-M^TM)_{ij},  \,\,  &\text{if} \,\,\ x_i \in \mathcal{N}_w(x_j) \,\, \text{or} \,\,  x_j \in \mathcal{N}_w(x_i)\\
0, \,\, &\text{otherwise.}
\end{cases}
\end{equation}	
Conversely, the between-class similarity matrix $\textbf{\emph{W}}_b$ depicts the statistical properties which are used to maximize the distance between classes and used as a high-order constraint. Distances between normal and pathological samples are computed as:
\begin{equation}\label{eq.simbetween}
W_b(i,j)=
\begin{cases}
1/k_2,  \,\,  &\text{if} \,\,x_i \in \mathcal{N}_b(x_j) \,\, \text{or} \,\,  x_j \in \mathcal{N}_b(x_i)\\
0, \,\, &\text{otherwise}
\end{cases}
\end{equation}	
with $\mathcal{N}_b$ containing $k_2$ neighbors having different class labels from the $i$th sample. The objective is to transform points to a new manifold $\mathcal{M}$ of dimensionality $d$, i.e. $x_i \rightarrow y_i$, by mapping connected tumor or normal samples in $\textbf{\emph{W}}_w$ as close as possible to the class cluster, while moving tumor and normal areas of $\textbf{\emph{W}}_b$ as far away from one another. This results in optimizing the following minimization and maximization objective functions:

\begin{equation}\label{eq.functions1}
    f_1=\min \dfrac{1}{2}\displaystyle \sum_{i,j}(y_i-y_j)^2  W_w(i,j) 
\end{equation}
    
\begin{equation}\label{eq.functions2}
   f_2=\max \dfrac{1}{2} \displaystyle\sum \limits_{i,j}(y_i-y_j)^2  W_b(i,j).
\end{equation}

\begin{figure}
\begin{center}
\includegraphics[width=1\linewidth]{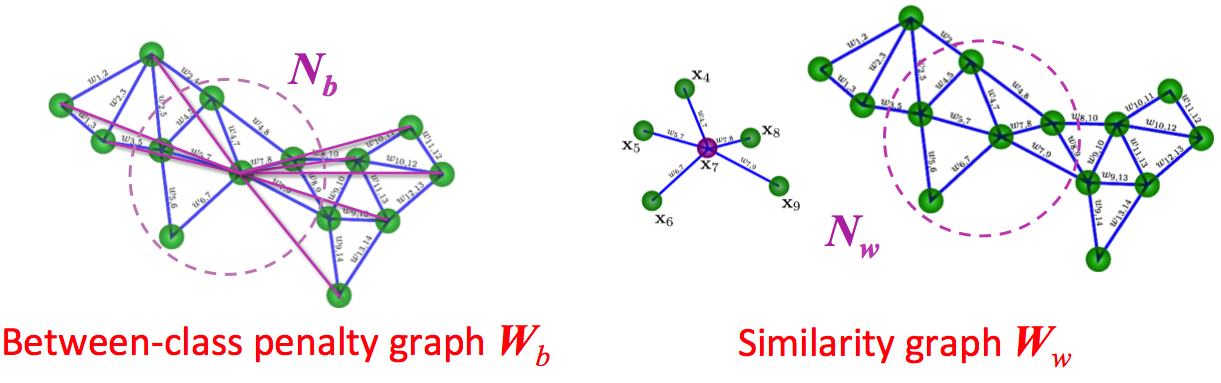}
  \caption{Illustration of the (left) between-class penalty graph $\textbf{\emph{W}}_b$  and (right) within similarity graph $\textbf{\emph{W}}_w$. The neighbourhoods of nodes between classes and within the same class composed of samples $x_i$ are defined by the terms $N_b$ and $N_w$, respectively.}
\label{fig.graphs}
\end{center}
\end{figure}

Here, Eq.(\ref{eq.functions1}) is used to penalize neighbouring points belonging to the same type if they are mapped far away within $\mathcal{M}$. Conversely, Eq.(\ref{eq.functions2}) penalizes embedded points of distinct classes, in the case where these are mapped close to each other within $\mathcal{M}$, as shown in Fig. \ref{fig.graphs}. Using the structure, it can be shown that an implicit representation of manifold points can be obtained by finding the similarity between points using the Grassmannian kernel $\mathbb{K}_{i,j}$, an appropriate mapping function can be obtained using  a supervised manifold learning approach, which will be described in the following section.

\subsubsection{Supervised manifold learning}

The optimal projection matrix, mapping new points to the manifold created in the training phase, is obtained by simultaneously maximizing class separability and preserving interclass manifold property, as described by the objective functions in Eqs.(\ref{eq.functions1})-(\ref{eq.functions2}). Assuming points on the manifold are known as similarity measures given by the Grassmannian kernel $\mathbb{K}_{i,j}$  defined in Eq.(\ref{eq.kernel}), a linear solution can be defined, i.e., $y_i=(\langle \alpha_1,x_i \rangle,\dots,\langle \alpha_r, x_i \rangle)^T$ for the $r$ largest eigenvectors with $\alpha_i=\sum_{j=1}^N a_{ij} x_j$. Defining the coefficient $\textbf{\emph{A}}_l=(a_{l1},\dots,a_{lN})^T$ and kernel $\textbf{\emph{K}}_i=(k_{i1},\dots,k_{iN})^T$ vectors, the output can be described as $y_i = \langle \alpha_l,x_i \rangle = \textbf{\emph{A}}_l^T \textbf{\emph{K}}_i$. By replacing the linear solution in the minimization and maximization of the between- and within-class graphs, the optimal projection matrix $\mathbb{A}$ is acquired from the optimization of the function as proposed by Harandi  \cite{Harandi11}. The algorithm generates implicitly a mapping  $\mathbb{A}$ using the  points on the Grassmannian manifold. The discriminant mapping $\mathbb{A}$ will enable the conservation of the global geometrical structure of the inherent distribution. The minimization of Eq.(\ref{eq.functions1}) becomes:
\begin{align}
f_1=&\sum_i \textbf{\emph{A}}_i^T \textbf{\emph{K}}_i \textbf{\emph{K}}_i^T \textbf{\emph{A}}_i^T W_w(i,i) - \sum_{i,j} \textbf{\emph{A}}_i^T \textbf{\emph{K}}_j \textbf{\emph{K}}_i^T \textbf{\emph{A}}_i^T W_w(i,j) \nonumber \\
=&\mathbb{A}^T \mathbb{K} \textbf{\emph{D}}_w \mathbb{K}^T \mathbb{A} - \mathbb{A}^T \mathbb{K} \textbf{\emph{W}}_w \mathbb{K}^T \mathbb{A}\\
=&\mathbb{A}^T \mathbb{K} \textbf{\emph{L}}_w \mathbb{K}^T \mathbb{A} \nonumber
\end{align}
given Eq. (\ref{eq.llegraph}). Here, $\mathbb{A}=[\textbf{\emph{A}}_1|\textbf{\emph{A}}_2|\dots|\textbf{\emph{A}}_r]$ and $\mathbb{K}=[\textbf{\emph{K}}_1|\textbf{\emph{K}}_2|\dots|\textbf{\emph{K}}_N]$. Similarly, the maximization of the between-class graph is defined as $
f_2=\mathbb{A}^T \mathbb{K} \textbf{\emph{L}}_b \mathbb{K}^T \mathbb{A}$, with $\textbf{\emph{L}}_b=\textbf{\emph{D}}_b-\textbf{\emph{W}}_b$. By combining $f_1$ and $f_2$, the optimal projection matrix $\mathbb{A}^*$ is acquired from the optimization of the function:
\begin{equation}
\mathbb{A}^* = \underset{\mathbb{A}}{\operatorname{argmax}} \, \mathbb{A}^T \mathbb{K} \textbf{\emph{L}}_b \mathbb{K}^T \mathbb{A} - h\mathbb{A}^T \mathbb{K} \textbf{\emph{L}}_w \mathbb{K}^T \mathbb{A}
\end{equation}
which is obtained by the eigenvalue decomposition method and $h$ is a Lagrangian multiplier that regularizes the final mapping function. Hence for any test point $x_q$, a manifold representation $v_q=\mathbb{A}^T \textbf{\emph{K}}_{q}$ is obtained using the kernel function based on $x_q$ and mapping $\mathbb{A}$.

\subsection{Fully-Connected Graph Segmentation}

Once the Grassmannian manifolds are obtained from the training phase, the segmentation problem is performed using a higher-order CRF model where clique potentials are inferred from the low-dimensional embeddings. In this sub-section, we describe the segmentation framework with the minimization of the energy term used in the process, as well as the definition of the energy terms defined in the manifold space.
\\
\\
We use higher-order conditional random fields \cite{Kohli09,Bauer11}, where instead of using direct voxel intensities which is a highly dimensional problem, clique potentials are derived from the trained discriminant embeddings in order to perform an unsupervised tumor segmentation. Potentials describing smoothness terms restrain traditional techniques in segmenting small structures in specific regions of interest, such as hyper-vascular tumors and irregularly shaped metastases. In fact in previous works, models generated using regular CRFs based on pairwise potentials yield very smooth results which do not follow the actual region of interest. Instead of using direct pixel intensities, we propose to employ potential measures obtained from the learned models to incorporate regional consistency in the segmentation process. A voting scheme will help to drive the delineation step as a smooth constraint.
\\
\\
We propose to incorporate manifold potentials defined in the discriminative high-dimensional feature space into the typical CRF model. The well known Gibbs formulation of this higher order CRF is adapted for this purpose and defined as:

\begin{equation}\label{eq.objfun}
   M(\mathcal{C}|\mathcal{X}) = \sum_{v_i \in \mathcal{X}} \psi(c_i | v_i)+\sum_{(v_i,v_j)\in \mathcal{E}} \phi(c_i, c_j | v_i, v_j) + \sum_{s \in \mathcal{S}} \xi_s(c_s | \textbf{{v}}_s).
\end{equation}

The random variables $\mathcal{X}$ denotes whether labels belong to the object of interest (tumor or not) in label space $\mathcal{C}$, $\mathcal{E}$ denotes edges connecting pairs of nodes and $\mathcal{S}$ denotes the set of segments, composed of multiple embedded patches. The CRF model defines the energy function based on the sum of unary $\psi$ and pairwise potentials $\phi$, as well as higher-order functions $\xi$, which will be described in the following section, where $v_i$ denotes the manifold projection of data point $x_i$ and $c_i$ is the label (tumor or normal) of node $v_i$. The nodes in the CRF model corresponds to the manifold embedded segments of $x_i$. Here, image segments $x_i$ are obtained from superpixels on each 2D slice, which consists of a normalized-cut segmentation method, followed by an iterative k-means clustering to create an image parcelled in equal-sized patches (average size of 12 pixels) that do not overlap \cite{Mori05}. Then, segments are mapped using the projection matrix such that $v_i=\mathbb{A}^T \textbf{\emph{K}}_{i}$ . Features are selected implicitly by the manifold from the low-dimensional space to obtain the best discrimination between the classes. Here, we utilize a fully-connected CRF model using a mean-field approximation of the original CRF such that the distribution is composed of a set of independent marginals minimizing KL-divergence. 

\subsubsection{Manifold-based potentials}
Here, we describe the three potentials that constitute the objective function of the CRF defined in Eq.(\ref{eq.objfun}).
\\
\\
\emph{Unary potentials} in the CRF model, represented by the term $\psi(v_i)$ of the energy function, represents the likelihood to assign a patch $i$ with a specific label. Typically, the unary potential is determined by the grayscale value of the voxel and the appearance model for each region of interest. However, intensity alone is not sufficient to discriminate between similar groups and often produces erroneous tumor delineations. Instead, we propose to utilize elaborate potential functions based on manifold representations of the classifier $\psi(c_i | v_i)=-\log(P(c_i | v_i))$, which evaluates the likelihood that the extracted patch belongs to the region of interest based on its embedded location within the Grassmannian manifold. The likelihood is calculated based on the negative value of the log function.
\\
\\
\emph{Pairwise potentials} are expressed as non-parametric manifold dissimilarities, extending the Gaussian kernel formulation in feature space which uses mean-field approximations of fully-connected CRF models  \cite{Krahenbuhl11}. Instead of forcing Euclidean features to fulfill this task, pairwise potentials are conditioned on the input data such that $\phi(v_i, v_j)=\mu(v_i, v_j)\exp (-\textrm{d}(i,i,\mathcal{X},\mathcal{M}))$, where $\mu$ describes how compatible labels are assigned between nodes $v_i$ and $v_j$. The distance between points $i$ and $j$, under label $l$, is the conditional distribution of the label $l$, with:

\begin{equation}\label{eq.kernelcc}
    \exp (-\textrm{d}(i,i,\mathcal{X},\mathcal{M}))=P(v_j=l | v_i=l,\mathcal{X},\mathcal{M}).
\end{equation}

Assuming the Frobenius distance in manifold space can offer a non-parametric estimation of the dissimilarity measure, the pairwise potential can be defined by imposing a range $\sigma_f$ over which valuable information can be inferred when applying a Gaussian window:

\begin{equation}\label{eq.kernelcc}
   \phi(c_i, c_j | v_i, v_j) = w \exp \left(\dfrac{\| v_i - v_j \|^2_F}{2 \sigma_f^{2}} \right)
\end{equation}

with the parameter $w$ weighting the pairwise relations. 
\\
\\
\emph{Higher-order potentials} are quality sensitive functions that define the label inconsistency in regions (different labels are assigned to a set of neighboring segments), by adding edges between nodes that are not immediate neighbours. We use the strategy in \cite{Vineet12} where for a given clique $\textbf{{v}}_s$ grouping $t$ segments, $t$ different embeddings are generated with the manifold projection matrix $\mathbb{A}$. The variance of the embedded coordinates of $v_s$ for all $t$ data points is the used as a quality measure $Q(s):s \rightarrow \mathbb{R}$ for all consistent segments in $\textbf{{v}}_s$. The higher-order potential is defined as 
$\xi_s(\textbf{{v}}_s)= C(\textbf{{v}}_s)\frac{1}{T}\lambda_{q}$ when $C(\textbf{{v}}_s)$, which counts the number of segments in $\textbf{{v}}_s$ not taking the dominant label $s$,  is lower than a threshold $T$. In the case the count is over $T$, then $\xi_s(\textbf{{v}}_s)=\lambda_q$, where $\lambda_{q}$ produces a penalizing term using $Q(s)$, which defines the flexibility of the clique potential. This term will favour joining similar segments together, while penalizing other segments which are less likely to constitute a single region. As proposed by \cite{Kohli09}, the penalty term $\lambda_q$  measures how the output unary potential varies on all voxels of a segment in order to assess the reliability of the potential. Here, the model parameters which are assigned to the potentials are obtained using cross-validation on the training data.

\subsubsection{Energy minimization}

We minimize the higher order CRF defined earlier in Eq.(\ref{eq.objfun}) by decomposing the problem into sub-modular terms. Kohli et al. \cite{Kohli09} showed that robust higher order functions can be found by expanding or replacing the clique function using auxiliary binary terms. Optimal displacements are calculated for these algorithms in order to minimize the clique potentials. Therefore, the high-order displacement terms are transformed into sub-modular quadratic functions, and are optimized based on graph cuts.  The minima of the energy function represents the solution for the tumor segmentation based on a labelling process which assigns a class at voxel of the image, generating tumor and normal tissue segmentations.
\\
\\
Finally we apply a Primal-Dual algorithm called FastPD \cite{KomodakisPT11} which can efficiently solve the problem in a discrete domain by formulating the duality theory in linear programming. Compared to other methods which do not provide optimal solutions or require long computational times to approximate the global minimum, the advantage of FastPD lies in its generality and efficient computational speed. It also guarantees that the generated solution will be the best approximation of the true global optimum without the condition of linearity.
\\
\\
\subsection{Validation methodology}

\begin{center}
\begin{figure}
 \begin{minipage}[b]{0.49\linewidth}
  \centering
  \includegraphics[height=1.7in] {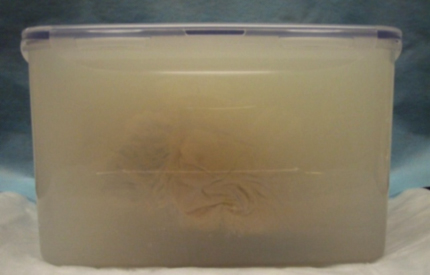}
  \centerline{(a)}
\end{minipage}
\begin{minipage}[b]{0.5\linewidth}
  \centering
  \includegraphics[height=2in] {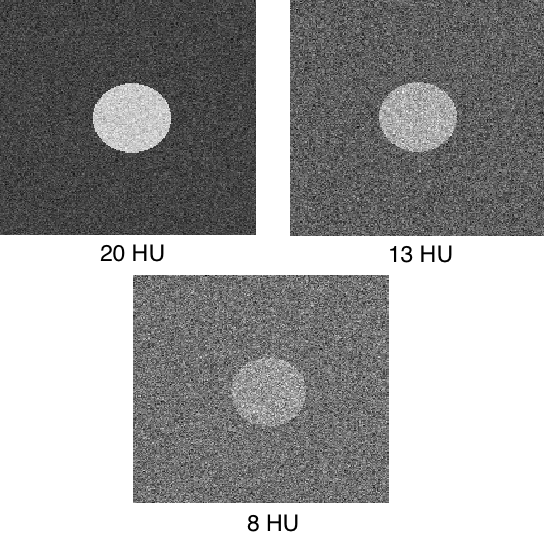}'
  \centerline{(b)}

\end{minipage}

    \caption{
      (a) Synthetic phantom using uniform agar gel (8\% agar w/v), replicating an abdominal region with a localized tumor. (b) Samples of artificial tumors at three different contrast levels (8, 13 and 20 HU).}
    \label{fig.phantom}
\end{figure}
\end{center}

To evaluate the performance of the algorithm in a controlled setting and help determine parameter values for the proposed framework, a dataset created from synthetic phantoms using uniform agar gel (8\% agar w/v) as shown in Fig. \ref{fig.phantom}(a), were used for an initial evaluation. Phantoms provide a medium to generate artificial data in a controlled settings, therefore providing gold-standard segmentation for reference. While they do not yield images similar to real patient images, it was shown in previous studies to provide significant information on the behavior of image processing algorithms \cite{Li13}. A total of nine phantoms were created, where each model included a tumor region composed of diluted dye agent in the middle of the phantom model. The size of the tumor (between 1 and 5cm), as well as the contrast of the tumor varied in each synthetic model. A CT image for each phantom was obtained with a 512 by 512 resolution and 1mm slice thickness. Gaussian noise with a s.d. of 1.5 was added to the images, simulating point granularity during image acquisition. This way, nine tumors were artificially simulated at different levels of contrast  (8, 13 and 20 Hounsfield units) (Fig. \ref{fig.phantom}(b)), with three different tumor diameters (1, 3 and 5 cm).
For the first clinical dataset, experienced radiologists generated ground-truth segmentations for the test data, and the evaluation was conducted by comparing the automated segmentations to the interobserver variations obtained by ground-truth segmentations. Measures such as  (1) volumetric overlap error (\%), (2) the average symmetric surface distance (mm), (3) the root mean square (RMS) symmetric surface distance (mm), (4) the relative absolute volume difference (\%), and (5) the maximum symmetry surface distance (mm) were computed for each segmentation based on the 2008 MICCAI liver tumor segmentation challenge  \cite{Deng08}. A score between 0 and 100 was assigned to each measure based on the scoring system designed for the tumor segmentation challenge to compute an overall score based on human segmentation. For a perfect segmentation, a metric will get a score of 100 when the value is an exact match (error=0). A reference point with a score 90 for each metric is determined from the segmentation performed by independent users. It represents the score a human observer can get by manual segmentation. The values of the five metrics for the reference point are as follows: (1) volumetric overlap error = 12.94 \%, (2) relative absolute volume difference = 9.64\%, (3) average surface distance = 0.40mm, (4) RMS distance = 0.72 mm, (5) maximum symmetry surface distance = 4.0 mm.  The score of each metric for a segmentation is then obtained using linear interpolation or extrapolation between the two points specified above (exact match and reference point).
\\
\\
To set similar conditions to the experiments from the MICCAI challenge, the operator could identify a coarse region of interest (120x120) around the tumor area in a single 2D axial slice, in order to allow a more narrow search region that surrounds the target tumor. The modified search area provides the method with a better sample tissue representation, which can help the discrimination process between normal and pathological tissue. The final segmentation obtained with the selected search region was used for comparison to the ground-truth segmentation.
\\
\\
For the second clinical dataset, the same five quantitative measurements defined previously were obtained by comparing the automated segmentations with the ground-truth manual segmentations, which were obtained by two independent experienced radiologists. No user interaction was required as the datasets were already processed to segment the liver from CT images \cite{zhang2010automatic}. The manifold learning and segmentation algorithm were programmed in C++, running on a 3.4 GHz Intel Core i7 CPU with 32 GB of RAM. The average training time took approximately 6h, which included the patch extraction and learning the manifolds offline.

\section{Results}
In this section, we present the experiments performed on both synthetic and clinical datasets of CT images with liver tumors. Results obtained from both qualitative and quantitative measures were compared to manual ground-truth segmentations, and confronted to state-of-the-art tumor segmentation approaches. 

\subsection{Artificial data}

\begin{center}
\begin{figure}
 \begin{minipage}[b]{0.99\linewidth}
  \centering
  \includegraphics[height=3in] {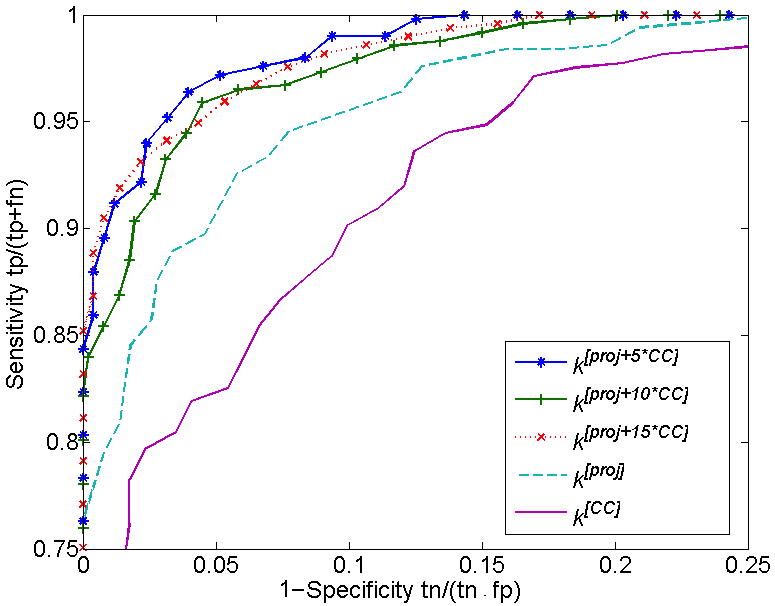}
  \centerline{(a)}
\end{minipage}
\begin{minipage}[b]{0.99\linewidth}
  \centering
  \includegraphics[height=3in] {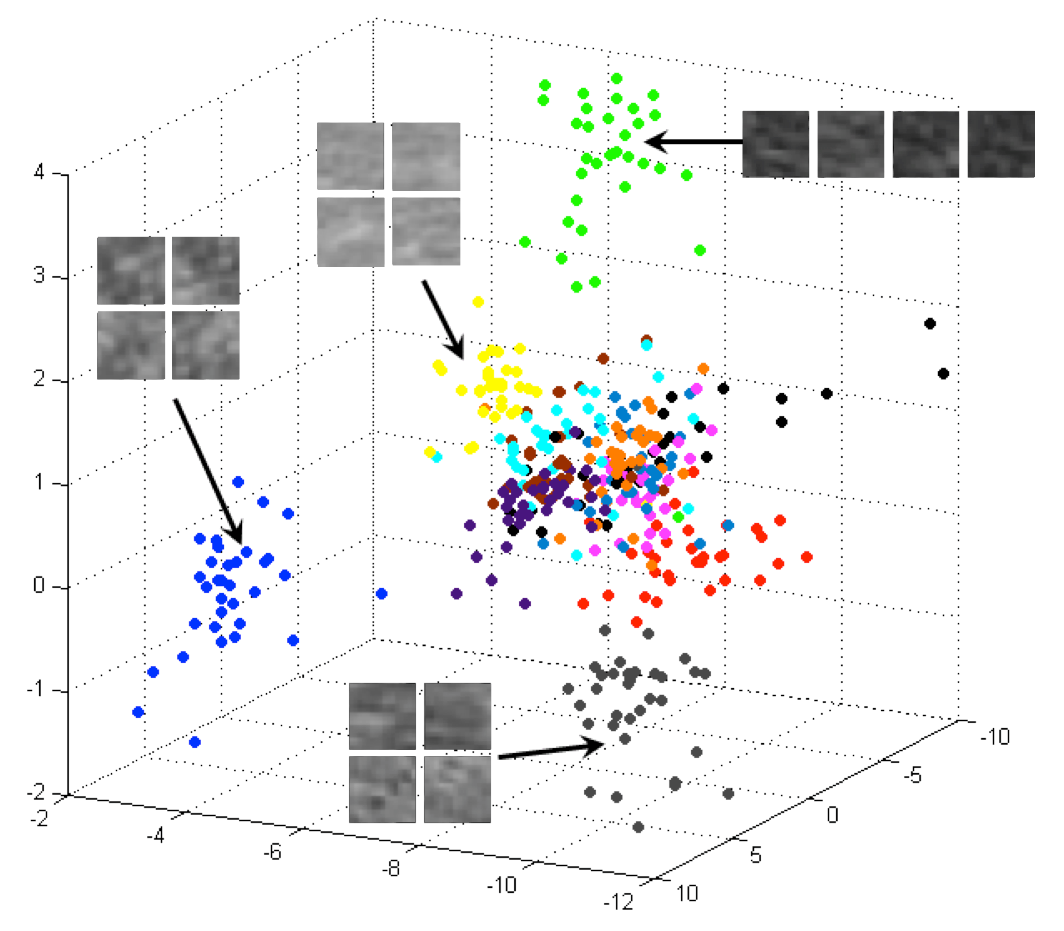}
  \centerline{(b)}

\end{minipage}

    \caption{
      (a) Comparison in ROC curve accuracy using 3 types of canonical correlation kernels. (b) Resulting 3D manifold embedding of testing images from normal liver tissue and pathological regions (blue, gray, green, yellow).
    }
    \label{fig:fig3}
\end{figure}
\end{center}

The Grassmannian manifold was trained with ground-truth datasets and the optimal neighborhood size was found at $k_1=10$ for within-class similarity graphs ($\mathcal{N}_w$), and $k_2=5$ for between-class neighborhoods ($\mathcal{N}_b$). The optimal manifold dimensionality was set at $d=6$, when the trend of the nonlinear residual reconstruction error curve stabilized for the entire training set. Fig. \ref{fig:fig3}a shows the ROC curves when using different types of kernels ($k^{[CC]}, k^{[proj]}, k^{[CC+proj]})$, illustrating the increased accuracy using the combined kernel ($\alpha_1=1, \alpha_2=5, h=1$), which suggests each are extracting complementary features from the training data. A 3D embedding of the computed manifold is represented in Fig. \ref{fig:fig3}b, showing the group of training image patches provided by the tumor and normal liver patches embedded in a single continuous manifold structure. 
\\
\\
Segmentation accuracy results where then obtained for all the synthetic datasets using a leave-one-out cross validation, studying the effect of contrast and tumor size from the segmentations of nine tumors. The results from this evaluation are presented in Table \ref{table:phantom}. Unsurprisingly, these show that the method's  performance in segmentation accuracy increases as the contrast level of the tumor increases as well. The accuracy with a contrast level of 8 Hounsfield units (HU) yields the lowest scores, while a contrast level of 20 HU generated the highest accuracy in segmentation. Typical levels of contrast for tumors in patients treated with RFA is of 10 HU. Results presented here can be used as a reliability indicator of the segmentation result of real tumors. Another factor which was evaluated in this experiment was the effect of the tumor size on the segmentation results. As the tumor tends to be larger, it provides the algorithm with increased contextual information that can be integrated  within the higher-order clique potentials. On the other hand, these artificial tumors are constituted of primarily homogenous regions, while real patient data often presents heterogeneous portions, which is typically more challenging.

\begin{table}[!t]
\caption{Results using artificial data generated from synthetic agar get phantoms, using three different contrast levels with 3 different diameter sizes. Quantitative measures were generated for each of the nine tumors, comparing results to ground-truth data.}
\label{table:phantom}
\renewcommand{\arraystretch}{1.3}
\centering
\scalebox{0.8}{
\begin{tabular}{l c c c c c}
\hline
  & Overlap & Vol. Diff.  &  Avg. Surf.  & RMS Surf. & Max. Surf. \\
  & error (\%) & (\%)  &  Dist. (mm)  & Dist. (mm) & Dist. (mm)\\
\hline
20 HU (1cm diam.) & 20.48 & 17.93 & 0.76 & 0.91  &  2.85 \\
13 HU (1cm diam.)&  24.17 & 21.04 &  0.99 & 1.38  & 3.42 \\
8 HU  (1cm diam.) &  34.84 & 29.11 & 1.25  & 1.77  & 3.80  \\
\hline
20 HU (3cm diam.) & 16.72 & 13.39  & 0.57 &  0.67  & 2.16  \\
13 HU (3cm diam.)& 21.82 & 17.01   & 0.80 & 0.92  &  2.83 \\
8 HU  (3cm diam.) & 32.77 & 26.85  & 1.02 & 1.33 & 3.41 \\
\hline
20 HU (5cm diam.) & 13.12  & 10.32 & 0.38 & 0.51 & 1.97 \\
13 HU (5cm diam.)&  18.55 & 14.24 & 0.56 & 0.78  & 2.69 \\
8 HU  (5cm diam.) & 28.45 & 22.37 &  0.76 & 1.03  & 3.10  \\
\hline
\end{tabular}
}
\end{table}
\subsection{Challenge dataset}

Table 2 presents the evaluation results from all 20 test cases provided by the tumor segmentation challenge dataset, while the distribution of scores compared to human raters is shown in the box-plots diagram in Fig. \ref{fig.boxplots}. In the evaluation, the training data was not included. The average time for segmenting a tumor was 53sec, given an input ROI. Segmentations were compared to an active contours approach with local Gaussian distributions \cite{Wang09} and to a texture classification method \cite{Bauer11}. The average and RMS surface distances ($1.4 \pm 0.3$ and  $1.6 \pm 0.4$mm) of the proposed method were significantly lower ($p \leq 0.05$) than distances generated by \cite{Wang09} and \cite{Bauer11}. Typical results of metastatic liver tumors segmented on CT images are shown in Fig. \ref{fig:challenge}. As it was observed in a previous study, the gallbladder represents an important cause of wrong classification of perihepatic metastasis detection since it exhibits similar intensity levels to many metastases, while being close to the location of the liver. The gallbladder can generate a convex-shaped indentation in the liver capsule, changing the morphology of the liver capsule.
\\
\\
The average score (using the challenge scoring system) for overlap error between automated and manual segmentation was $77.3 \pm 12.5$. Boxplots show the proposed method performs similarly to interobserver variability in all quantitative measurements. As demonstrated in \cite{Hame12},  segmentation results were not as accurate for larger tumors compared to smaller sized tumors. By analyzing  results for tumors with higher volumes,  the average score  of $59.1$ was obtained, while the  overlap error and volume error were similar to average results. On the other hand, surface distance errors were higher, with a average distance of 3.06mm. This can be explained by the less-uniform appearance of larger tumors, which may include hyper-vascular or necrotic regions. Furthermore, larger tumors are often located close to the liver edges, causing misclassification with surrounding organs as well.
\begin{center}
\begin{figure}
 \begin{minipage}[b]{0.32\linewidth}
  \centering
  \includegraphics[height=2.9in, width=1.6in] {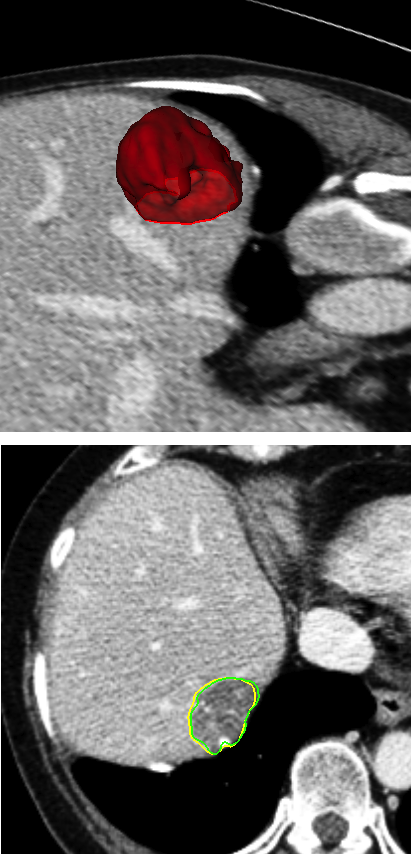}
    \centerline{(a)}
\end{minipage}
\begin{minipage}[b]{0.32\linewidth}
  \centering
  \includegraphics[height=2.9in, width=1.6in] {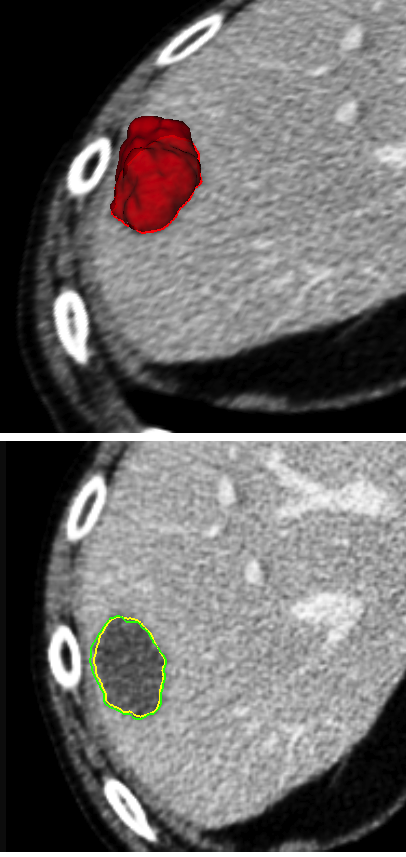}
    \centerline{(b)}
\end{minipage}
\begin{minipage}[b]{0.32\linewidth}
  \centering
  \includegraphics[height=2.9in] {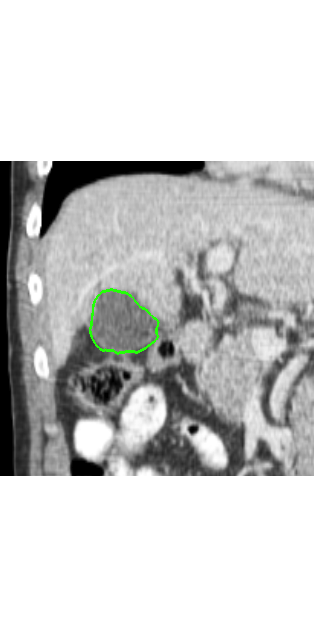}
    \centerline{(c)}
\end{minipage}
    \caption{
      (a)-(b) Example cases of segmentations on two cases from challenge dataset. Top row show the 3D segmented model. Bottom row show contours of automatic (green) and manual (yellow) segmentations. (c) Sample misclassification of the gallbladder. 
    }
    \label{fig:challenge}
\end{figure}
\begin{table}[!t]
\caption{Error metrics from the CT liver tumor segmentations on challenge dataset. We present results using only unary and pairwise ($\psi+\phi$) and unary, pairwise higher-order terms  ($\psi+\phi+\xi$).}
\label{table:CTres}
\renewcommand{\arraystretch}{1.3}
\centering
\scalebox{0.8}{
\begin{tabular}{l c c c c c}
\hline
  & Overlap & Vol. Diff.  &  Avg. Surf.  & RMS Surf. & Max. Surf. \\
  & error (\%) & (\%)  &  Dist. (mm)  & Dist. (mm) & Dist. (mm)\\
\hline
LGD \cite{Wang09} &   $ 33.1 \pm 2.9 $ &	 $23.6\pm 4.5 $ &	$ 2.3 \pm 0.5 $	&	$2.6 \pm 0.8 $ &	$ 8.9 \pm 2.0 $\\
\hline
SVM+CRF \cite{Bauer11} & $  31.5 \pm	2.7 $ &	$ 22.2 \pm 4.3 $ &	$2.0 \pm 0.5 $	&	$ 2.3  \pm 0.6$ &	$ 8.3  \pm 1.8 $\\
\hline
Proposed  $\psi+\phi$ (n=20) & 27.9 $\pm$ 2.1 & 19.5 $\pm$ 3.6  & 1.7 $\pm$ 0.6 & 2.0 $\pm$ 0.5 & 7.2 $\pm$ 1.9\\
Proposed $\psi+\phi+\xi$ (n=20) & \textbf{25.2 $\pm$1.7} & \textbf{14.3 $\pm$ 2.8}  & \textbf{1.4 $\pm$ 0.3} & \textbf{1.6 $\pm$ 0.4} & \textbf{6.9 $\pm$ 1.8}\\
\hline
\end{tabular}
}
\end{table}

\end{center}
\begin{figure}[t!]
  \centering
  \includegraphics[width=4.7in ] {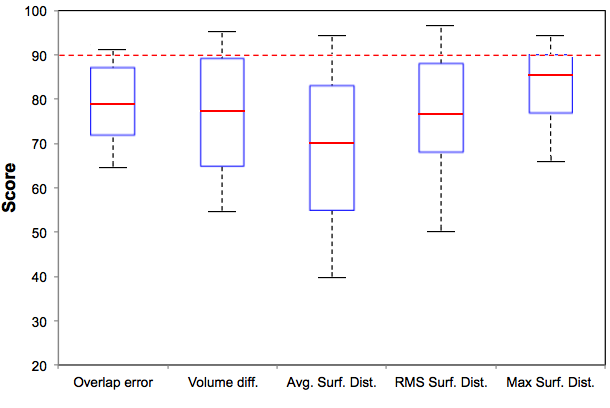}
\caption{Scores of the challenge dataset presented as a boxplot diagram. Score of average interobserver variability (90) is shown with dashed line for reference.}
\label{fig.boxplots}
\end{figure}

\subsection{Clinical dataset}
\begin{figure}[t!]
  \centering
  \includegraphics[width=4.5in] {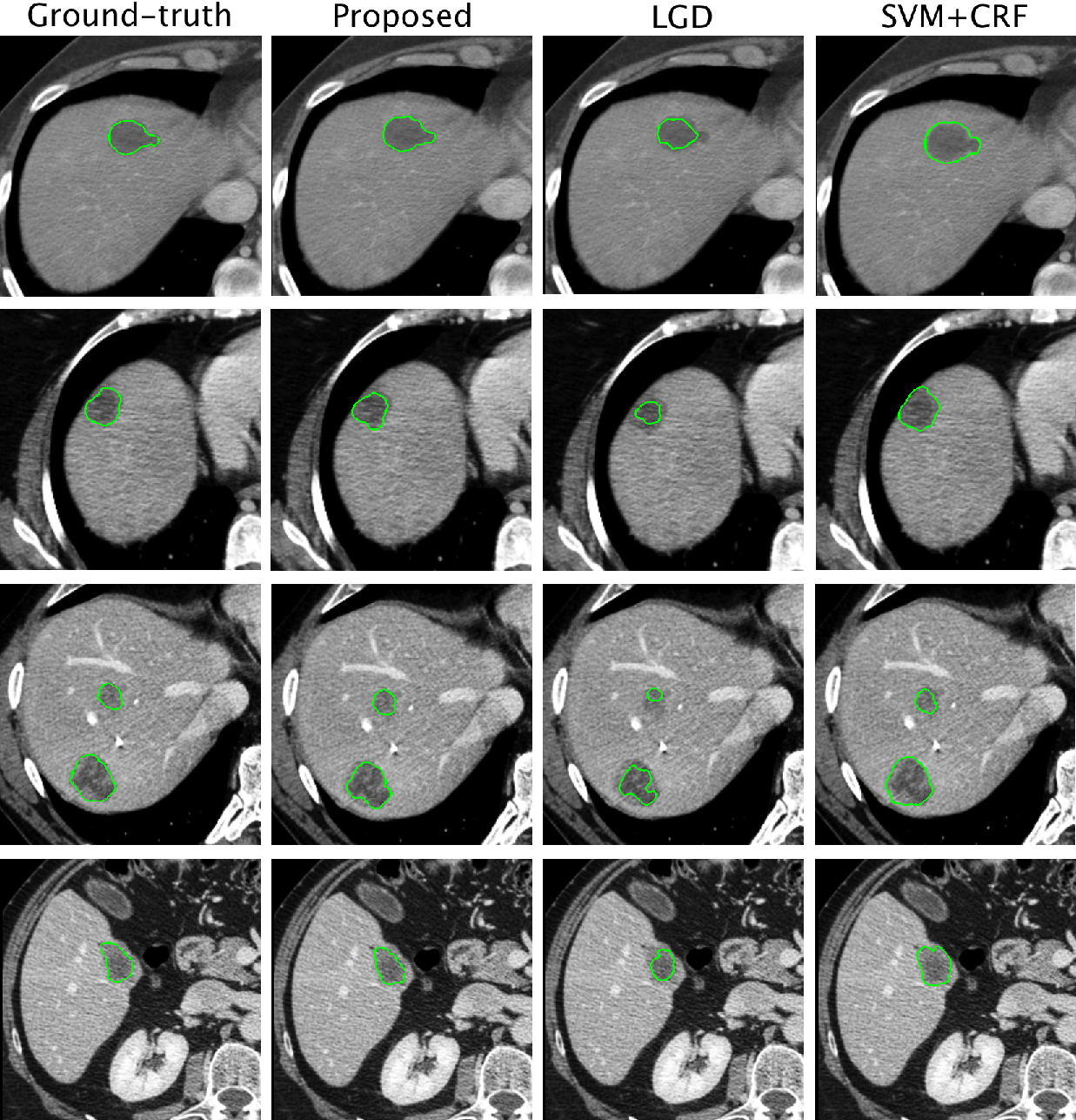}
\caption{Sample segmentation results of metastatic liver tumors from four different cases in the clinical dataset. First column shows ground-truth delineation verified by a radiologist, the second column presents results obtained with the proposed segmentation approach, the third using Gaussian distributions \cite{Wang09} and the last column with texture classification with SVM's. }
\label{fig.CT}
\end{figure}

\begin{figure}[t!]
  \centering
  \includegraphics[width=4.6in ] {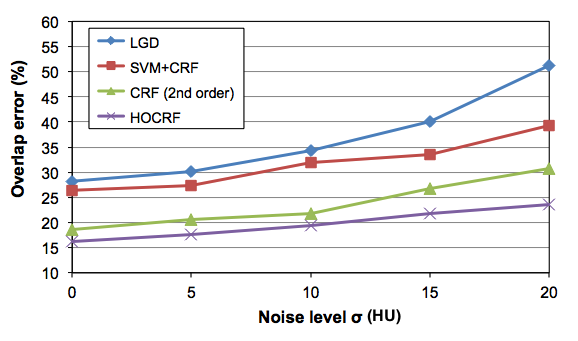}
\caption{Evolution of overlap error (\%) with increasing Gaussian noise levels added to input images. Results show the increased robustness of higher-order graphical models (HOCRF) to traditional methods such as LGD and SVM.}
\label{fig.noise}
\end{figure}

\begin{figure}[t!]
  \centering
  \includegraphics[width=5in ] {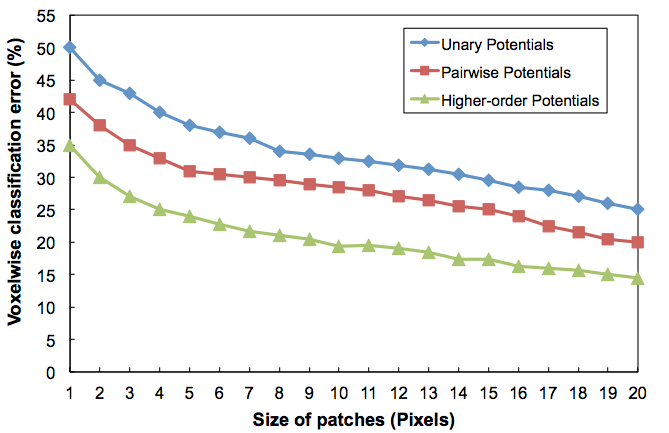}
\caption{Tumor voxel classification error in the proposed method. The figure illustrates how the global voxelwise classification rates changes as the patch sizes increase within the areas of interest.}
\label{fig.region}
\end{figure}

Table \ref{table:CTres} shows the performance of the method with 10 training and 30 test cases from the clinical dataset. The average segmentation time per tumor was of 1min 42sec, since the automatic process included the entire liver region. Results were also compared to an active contours approach with local Gaussian distributions \cite{Wang09} and to a texture classification method \cite{Bauer11}. The average and RMS surface distances ($0.7 \pm 0.4$ and $1.4 \pm 0.2$mm) of the proposed method were significantly lower ($p \leq 0.05$) to distance measures generated by \cite{Wang09} and \cite{Bauer11}. In terms of overlap error, the proposed approach reduces this error by $10\%$ compared to the top ranking texture classification. Typical problems occurred in the periphery of the tumors and in cases of rim-enhancing liver metastases. These cases offer a density which was not observed in the training set, but could be compensated with additional data in the manifold. Fig. \ref{fig.CT} shows sample segmentation results.
\\
\\
By assessing the gain in accuracy when adding the higher-order terms in the energy formulation, the overlap error is reduced by 2.2$\%$, which is a statistically significant difference ($p \leq 0.05$) to the second-order MRF model. In order to evaluate the robustness of the method, we performed additional experiments by measuring segmentation accuracy with 4 different levels of Gaussian noise added to the input images. Fig. \ref{fig.noise} demonstrates that the proposed methodology possesses increased tolerance to noise compared to the other methods. Furthermore, the inclusion of the higher-order term improves the robustness to standard pairwise CRFs. Finally, segmentation accuracy was assessed by computing the amount of  voxels classified in the wrong group within the zones surrounding the tumor contours, instead of the entire volume. A comparison between different segmentation methods is shown in Fig. \ref{fig.region}. Hence, the evolution of the voxelwise classification errors decreases as the size of the evaluated patches around the tumor increases. The patch sizes did not affect computation times.

\begin{table}[!t]
\caption{Error metrics from the CT liver tumor segmentations on clinical dataset. We present results using only unary and pairwise ($\psi+\phi$) and unary, pairwise higher-order terms  ($\psi+\phi+\xi$).}
\label{table:CTres}
\renewcommand{\arraystretch}{1.3}
\centering
\scalebox{0.8}{
\begin{tabular}{l c c c c c}
\hline
  & Overlap & Vol. Diff.  &  Avg. Surf.  & RMS Surf. & Max. Surf. \\
  & error (\%) & (\%)  &  Dist. (mm)  & Dist. (mm) & Dist. (mm)\\
\hline
LGD \cite{Wang09} &   $ 28.3 \pm 3.5$ &	 $19.7\pm 5.5 $ &	$ 1.4 \pm 0.4 $	&	$2.0 \pm 0.7 $ &	$ 8.2 \pm 2.2 $\\
\hline
SVM+CRF \cite{Bauer11} & $  26.6 \pm	3.0 $ &	$ 16.1 \pm 4.6 $ &	$1.2 \pm 0.4 $	&	$ 1.8  \pm 0.6$ &	$ 7.7  \pm 1.9 $\\
\hline
Proposed (Training n=10) & 2.0 $\pm$ 0.5 & 0.1 $\pm$ 0.1 & 0.1 $\pm$ 0.1 &  0.4 $\pm$ 0.2 & 1.1 $\pm$ 1.0 \\
Proposed $\psi+\phi$ (Testing n=30) & 18.6 $\pm$ 1.9 & 12.4 $\pm$ 2.9  & 0.9 $\pm$ 0.3 & 1.5 $\pm$ 0.4 & 6.8 $\pm$ 1.9\\
Proposed  $\psi+\phi+\xi$ (Testing n=30) & \textbf{16.4 $\pm$1.7} & \textbf{9.9 $\pm$ 2.3}  & \textbf{0.7 $\pm$ 0.4} & \textbf{1.4 $\pm$ 0.2} & \textbf{6.1 $\pm$ 1.6}\\
\hline
\end{tabular}
}
\end{table}

\section{Discussion}
\label{sec:discussion}

We proposed a new, accurate and adaptable method for liver tumor segmentation. This was achieved through a higher-order fully-connected graphical model that was optimized using potential functions defined in a discriminant Grassmannian manifold. This increased the ability to isolate diseased liver regions from normal tissue as compared to state of the art CRF or SVM techniques. A thorough validation on subjects with metastatic liver tumors was undertaken to evaluate the performance of the method, leading to promising results. Based on the comparative measurements to ground-truth delineations, the resulting segmentations showed to be accurate for metastatic liver tumors, while yielding similar results to interobserver variability. The method's performance was tested on two distinct collections of tumors. For the clinical dataset provided by two hospitals, the proposed framework achieved slightly better results than other methods used for liver tumor segmentation. The average overlap error was improved compared to recently published method using local Gaussian distribution fitting. The evaluation with the challenge dataset illustrated the fact the approach yielded encouraging results, even on images with little contrast, noisy data or in cases where the tumor boundaries were not as clear.
\\
\\
The boundaries of the tumors represent the greatest source of discrepancy between manual and automated methods, as it remains difficult to set hard limits for the segmentation area. In some case, undersegmentation was present when the transition in voxel intensities between tumor and normal liver tissue deviates significantly from the learned manifold. In this case, the discriminant framework would classify these points as normal since the projection matrix tends to map unknown sample points closer to the physiological normal class. On the other hand, in the experiments presented in this paper, the manifolds were representative of the data distribution in order to model intensity variations that were present in the testing dataset. While we did not experiment testing the method on the clinical dataset using a model trained on the challenge dataset, it is not expected to penalize the performance of the segmentation algorithm as both datasets are similar in tumor type and imaging modality. However testing on non-enhanced CT images using manifold embeddings trained on contrast-enhanced CT is a current limitation. The method was also able to handle various shapes of liver tumors, such as ellipsoid or elongated shapes. Another source of possible difference is the distinction between necrotic and active tumor regions, which produces a greater level variability among raters in the challenge dataset.
\\
\\
The method relies on a set of parameters which may affect the performance of the algorithm, such as the intrinsic dimension of the discriminant manifold. A careful choice for this parameter can significantly optimize the performance of the method. Our experiments chose this parameter carefully by selecting the dimension yielding the lowest residual variance in reconstruction error. The within and in-between similarity graphs also relies on the size of their respective neighbourhoods, which affects the compactness or spread of the points on the manifold. Another set of important parameters are the weights applied to combined kernel functions, which controls the importance of both the projection and canonical correlation kernels. These kernels are often application specific and can be determined once during the training phase. A limitation of the method is the relatively high computational cost for learning the discriminant manifold embeddings during the learning process, which can take a few hours of computation. However this process is performed offline and can potentially be parallelized.  
\\
\\
In the datasets used in this study, liver metastases were a result of primary colorectal cancer. The results for lesions mainly caused by other types of cancer might differ. The evaluation on the multiple raters segmenting the data has shown that, overall, the proposed method significantly reduces both inter- and intra-observer variability of volumetric measurements. This introduces a certain level coherence in the delineation of liver metastases, and can be of significant value for the application of the RECIST criterion, in order to obtain repeatable results through multiple patient scans. For liver metastases, a limitation of the evaluation is that the intra-observer variability for the delineation of the tumor between raters can be higher to the difference between unsupervised and manual approaches. Another observation was that as the lesion increases in size, the variability between the segmentations is reduced, mainly because smaller differences in the segmentations have a lower relative influence.
\\
\\
An improvement of the approach will be to perform a series of experiments to determine the accuracy of the resulting model when providing a more localized initialization by automatically detect position of potential metastases with the segmented liver shape and limiting the number of potential false-positives generated by the algorithm. Furthermore, the model can be improved to enable the extraction of multiple sub-regions within the tumor (necrotic, active tumor), as well as include the capability of adaptation to unconventional tumor configurations, such as H-shaped tumors and more diffuse or infiltrating metastases. Integration of advanced graphical models in the local mesh adaptation process can potentially increase the delineation of detailed tumor boundaries. Translating the algorithm to other modalities, such as MRI and 3D ultrasound for interventional use, is also envisioned for interventional guidance of RFA procedures. Finally, a multi-site evaluation of this technique would be beneficial towards wide-spread clinical adoption.
\\
\\
The proposed framework based on discriminant manifolds is general, and can be extended to other applications in vision (segmentation) and medical imaging (Alzheimer's diagnosis, clinical decision support systems), to accommodate for modelling object tissue variations of similar shape and size, where disease detection can be of significant clinical value. The proposed method promises to facilitate and accelerate quantitative image analysis for clinical diagnosis with MRI or CT. Encoding prior knowledge relating to shape representation is a natural extension of the proposed formulation which can allow to ensure consistency in the final shape segmentation. Finally, the approach can also be extended to other pathologies such as glioblastoma multiforme by segmenting high-grade gliomas in MRI. In this case, the manifold potentials could be trained to characterize different necrotic, active and edema regions in the tumor from multi-parametric MRI, and help to automatically discriminate between normal and diseased brain tissue. Indeed, the framework easily allows to add multiple classes in the discriminant learning process, not only a two-class problem.  
\\
\\
\textbf{Acknowledgments}
The authors would like to acknowledge Dr. Nadine Abi-Jaoudeh from the National Institutes of Health for providing liver CT images. Research funding was supported in part by the Canada Research Chairs and the NSERC Discovery grant program and the Fonds ed recherchŽ du Quebec - Sant\'e (Cancer Award 26993). 
\\
\\
\textbf{References}
\bibliographystyle{elsarticle-num}
\bibliography{sample}







\end{document}